\documentclass[letterpaper, 10 pt, conference]{ieeeconf}
\usepackage{url}
\usepackage{textcomp}

\usepackage[font=scriptsize]{caption}

\usepackage{graphicx, multicol}
\usepackage[ruled,vlined]{algorithm2e}
\usepackage{tabulary}
\usepackage{color}
\usepackage{tikz, pgfplots}
\usepackage{tabularx}
\usepackage{amsmath, amssymb, amsthm}
\usepackage[
backend=biber,
style=ieee,
sorting=none
]{biblatex}
\IEEEoverridecommandlockouts                              

\AtBeginBibliography{\small}
\usepackage{mathtools}

\usepackage{comment}
\usepackage{caption}

\usetikzlibrary{calc,shapes.callouts,shapes.arrows}
\usepgfplotslibrary{fillbetween}
\pgfplotsset{compat=newest}
\usetikzlibrary{calc}
\usetikzlibrary{positioning}
\usepackage{caption}
\usepackage{subcaption}

\usetikzlibrary{external}
\definecolor{mycolor1}{RGB}{230,97,1}

\begin{document}
\title{\LARGE \bf Leveraging Heterogeneous Capabilities in Multi-Agent Systems for Environmental Conflict Resolution}

\author{Michael E. Cao$^1$, Xinpei Ni$^2$, Jonas Warnke$^2$, Yunhai Han$^2$, Samuel Coogan$^1$, and Ye Zhao$^2$
\thanks{This research was supported in part by the National Science Foundation under the National Robotics Initiative, award \#1924978. The authors would also like to thank Aziz Shamsah and Yuki Yoshinaga for their assistance in rendering the 3D simulation environment.}
\thanks{$^{1}$M. E. Cao and S. Coogan are with the School of Electrical and Computer Engineering, College of Engineering, Georgia Institute of Technology,
        Atlanta, GA 30332, USA.
        {\tt\small \{mcao34, sam.coogan\}@gatech.edu}}%
\thanks{$^2$J. Warnke, Y. Han, X. Ni and Y. Zhao are with the George W. Woodruff School of Mechanical Engineering, College of Engineering, 
        Georgia Institute of Technology, Atlanta, GA 30332, USA. 
        {\tt\small \{jwarnke3, xni32, yhan389, yzhao301\}@gatech.edu}.}%
\thanks{The authors would like to thank Prof. Seth Hutchinson's support on the Digit humanoid robot hardware and Profs. Aaron Young and Greg Sawicki for their support on experimental space and motion capture facility.}
}

\maketitle

\begin{abstract}
In this paper, we introduce a high-level controller synthesis framework that enables teams of heterogeneous agents to assist each other in resolving environmental conflicts that appear at runtime. This conflict resolution method is built upon temporal-logic-based reactive synthesis to guarantee safety and task completion under specific environment assumptions. In heterogeneous multi-agent systems, every agent is expected to complete its own tasks in service of a global team objective. 
However, at runtime, an agent may encounter un-modeled obstacles (e.g., doors or walls) that prevent it from achieving its own task. To address this problem, we employ the capabilities of other heterogeneous agents to resolve the obstacle. 
A controller framework is proposed to redirect agents with the capability of resolving the appropriate obstacles to the required target when such a situation is detected.
Three case studies involving a bipedal robot Digit and a quadcopter are used to evaluate the controller performance in action. Additionally, we implement the proposed framework on a physical multi-agent robotic system to demonstrate its viability for real world applications. 
\end{abstract}

\section{Introduction}
\label{section:introduction}
\begin{figure*}[ht]
\begin{subfigure}[t]{.32\textwidth}
  \centering
  \includegraphics[width=0.9\linewidth]{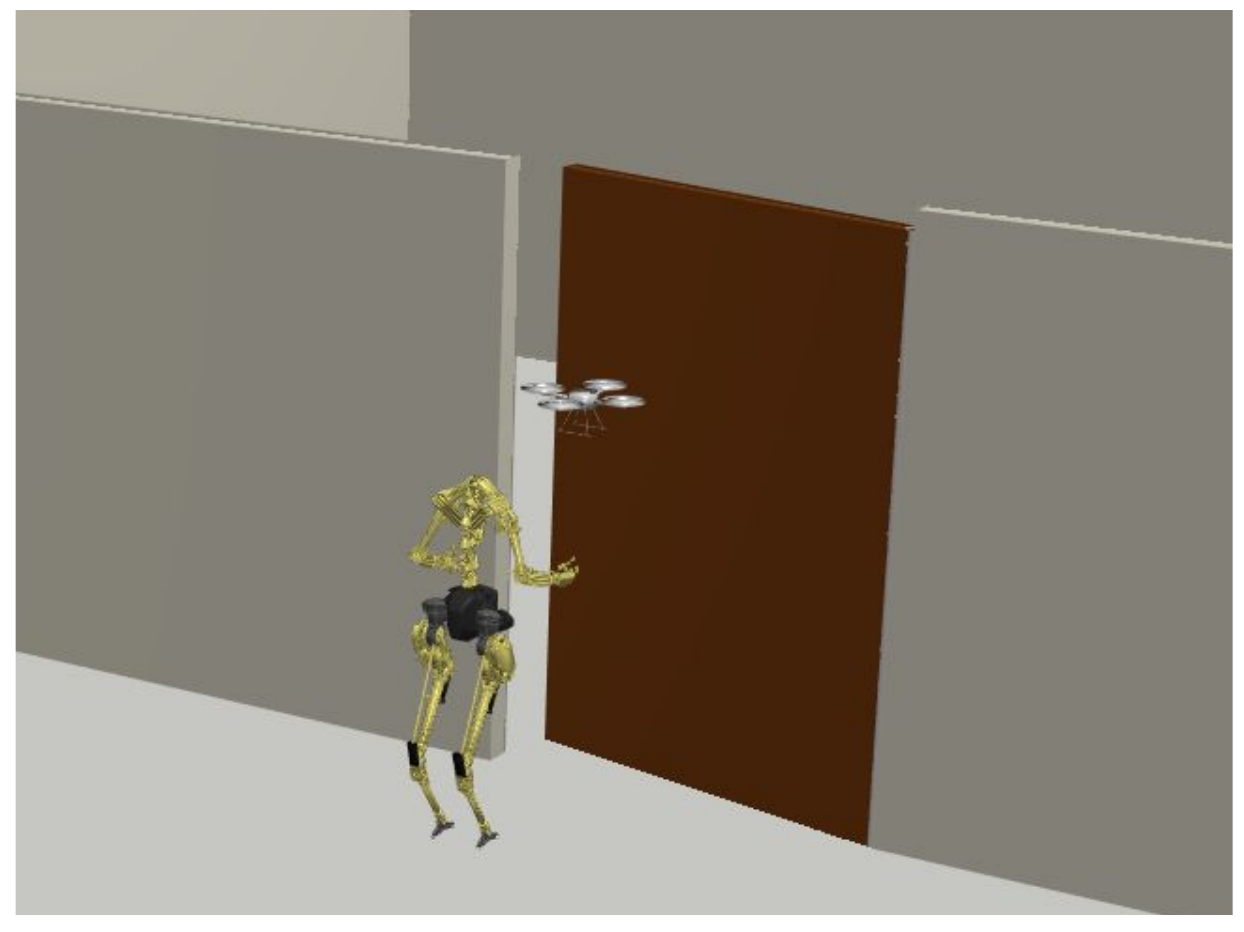}
  \caption{Simulation set-up with a bipedal robot opening a door for the quadcopter}
  \label{fig:Sim_sub1}
\end{subfigure}
\hfill
\begin{subfigure}[t]{.32\textwidth}
  \centering
  \includegraphics[width=0.9\linewidth, height = 2.8cm]{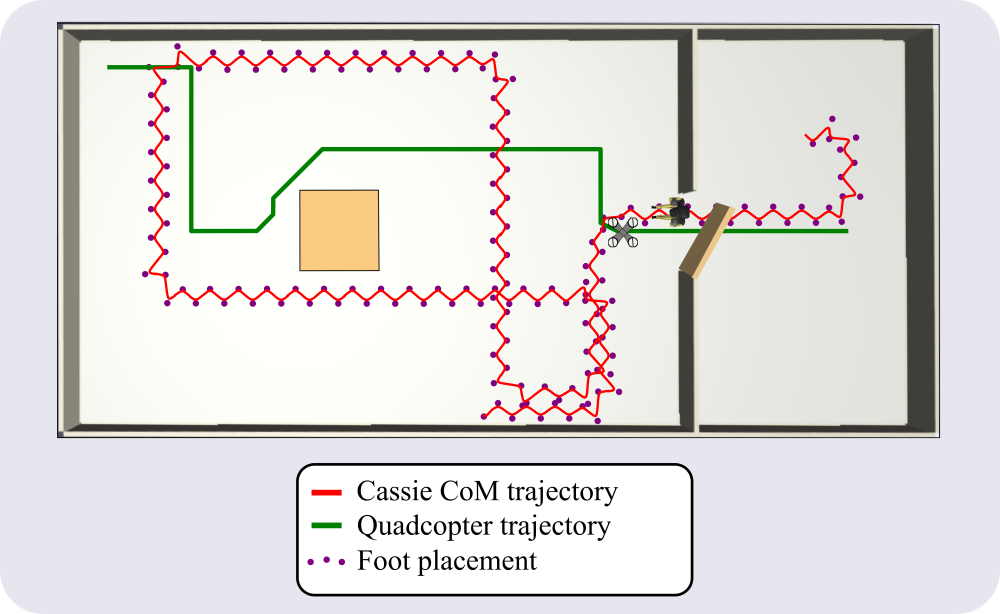}
  \caption{A bird's eye view of Digit's and the quadcopter's trajectories}
  \label{fig:Sim_sub2}
\end{subfigure}
\hfill
\begin{subfigure}[t]{.32\textwidth}
  \centering
  \includegraphics[width=0.9\linewidth]{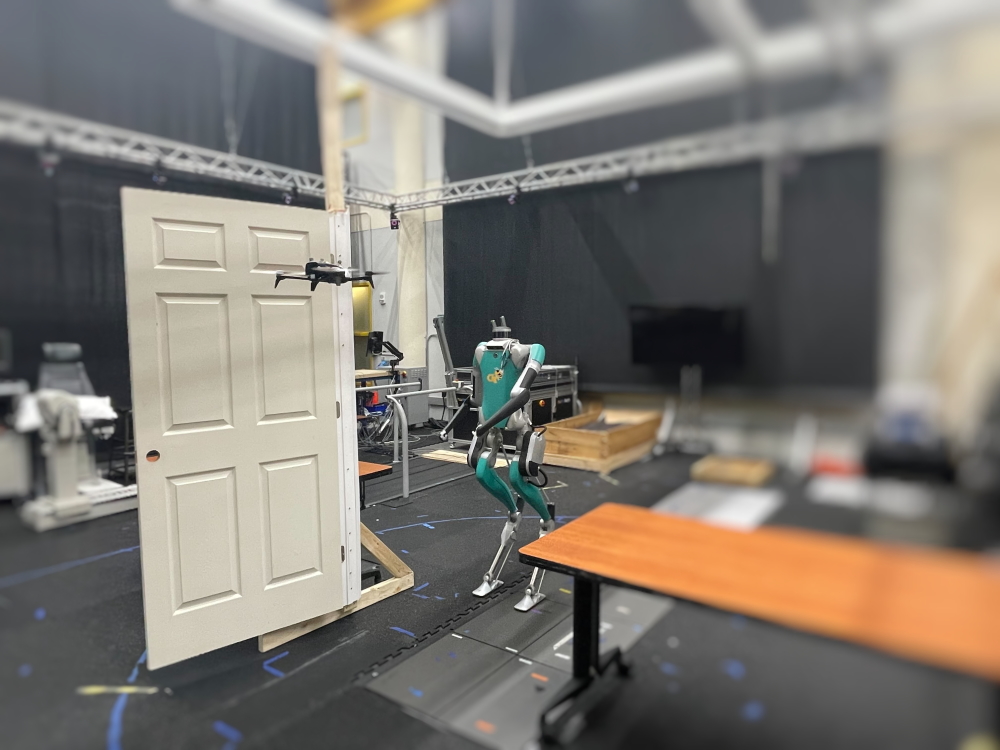}
  \caption{Hardware set-up with a Digit walking robot and a quadcopter}
  \label{fig:hardware_digit}
\end{subfigure}
\hfill
\setlength{\belowcaptionskip}{-0.5cm}
\caption{3D rendering of case study 1 in Drake simulation \cite{drake}. The quadcopter approaches an unknown door obstacle preventing it from achieving its goal. The quadcopter then requests assistance from a bipedal walking robot Digit. Accordingly, Digit opens the door and both agents resume their original task. As the formulation is built on a correct-by-construction controller, the actions that Digit takes are guaranteed to be safe (barring detection of another unexpected obstacle), though they may not take the most direct path available. The controllers are then implemented in hardware experiments.}
\label{fig:Sim}
\end{figure*}

Heterogeneous multi-agent systems with distinct mobility capabilities are generally capable of accommodating a larger variety of tasks than those consisting of a homogeneous team of agents~\cite{emam_adaptive, kit_exploration, Ziyi2022MultiRobot}. To achieve autonomous team behaviors such as the multi-room patrolling shown in Figure~\ref{fig:Sim_sub2}, a common approach is to automatically synthesize a controller for each agent, as this is often more efficient than crafting controllers by hand. However, creating controllers in this way has its own set of challenges, among which ensuring that the generated controllers do not cause any agents to perform tasks that would induce breakage or otherwise risk the agent's safety is a top priority~\cite{wong_synthesis}.

In this paper, we study controller synthesis to resolve environmental conflicts such that multi-agent team specifications are fulfilled using the generalized reactivity (1) (GR(1)) fragment \cite{gr1} of Linear Temporal Logic (LTL). The GR(1) fragment, in particular, allows for reactive synthesis algorithms that have favorable polynomial time complexity while retaining the ability to encode a large variety of specifications~\cite{liu2013synthesis, TL_Planning, gu2022pushrecovery}. These reactive synthesis methods 
are powerful because they provide formal guarantees on correct-by-construction controller behavior under any modeled action from the environment~\cite{reactiveSynth}.

A major challenge of reactive synthesis methods is that they require an explicit model of the environment's capabilities and may not be robust to unexpected changes in these capabilities encountered at runtime \cite{Wong2014CorrectHR}. For example, an unmodeled obstacle that interferes with the operation of the system may unexpectedly appear. Ensuring that the system's operation is robust to environmental changes is therefore an important area of research. To this end,  multiple lines of research have been proposed in the literature, such as online synthesis of local strategies which are further patched to the original controller \cite{livingston2012backtracking}, offline analysis of counter-strategies to resolve unrealizable specifications \cite{li2011mining}, controller synthesis that can tolerate a finite sequence of environmental assumption violations \cite{ehlers2014resilience}, or robust metric automata designed such that the system state is maintained within a bounded $\epsilon$-distance from the nominal state under unmodeled disturbances \cite{majumdar2011robust}. There also exist robustness methods that allow the system to identify specific broken environment assumptions~\cite{RamanAutofeedback}. However, none of these works studied the strategy of employing other agents to resolve environmental conflicts, which will be the focus of this paper, as this direction is underexplored. For example, the authors in~\cite{wong2018resilient} have studied the correction of broken assumptions, but focus on the cases where the broken assumption is due to unexpected behaviors by another agent operating within the workspace, which is resolved by changing that agent's behavior.

In heterogeneous teams, certain agents may have the capability to manipulate and correct a broken environment assumption for another agent. There have been works that explore the reconfiguration of the environment by the active agent in order to enable completion of its own objective~\cite{vandenBerg2010, Krontiris-RSS-15, vasilopoulos2021technical}. these works, however, did not focus on employing the assistance of other agents. Motivated by this, our study focuses on leveraging each agent's individual capability to resolve broken environment assumptions that prevent another agent from achieving its objective. Formally, we consider scenarios where the broken environment assumption renders an agent's specification unrealizable, yet another agent has the ability to fix the violation.

The contribution of this paper is to (i) propose a navigation planning framework enabling agents in a multiagent team to leverage their heterogeneous capabilities to assist each other in resolving environmental obstacles using reactive synthesis; (ii) implement the proposed framework on a physical multi-agent robotic system composed of a Digit~\cite{agility} bipedal robot  and a Parrot Bebop 2 quadcopter. This is the first time that the LTL-based planning method is implemented on a heterogeneous robot team including a full-size humanoid robot for navigation tasks and obstacle resolution. The proposed framework has the following four main components:
\begin{itemize}
    \item \textbf{Environment Characterization:} Observe the environment at runtime and verify whether the next state in the controller automaton would satisfy or violate any new specifications generated from runtime observations.
    \item \textbf{Safe Action Replanning:} 
    Backtrack states in the automaton and replace an action that will lead to a safety violation with a known safe maneuver.
    \item \textbf{Violation Resolution:} Identify other agents with the capability of resolving the violation and assign one to be responsible for violation resolution.
    \item \textbf{Task Replanning:} Add the resolution of the violation to the assigned agent's objectives and trigger a change in behavior of that agent to execute its new objective.
\end{itemize}
We refer to these components collectively as the ``coordination layer" that interacts with the other elements of the controller (see Figure~\ref{fig:frame}).

The rest of the paper is outlined as follows: In Section~\ref{section:preliminaries}, we introduce the basics of LTL specifications and GR(1) formulas. We then formally define our problem statement in Section~\ref{section:problem formulation}. In Section~\ref{section:controller synthesis}, we provide an overview of the previous work that this study is built upon before detailing the main approach that enables heterogeneous cooperation in Section~\ref{section:approach}. Section~\ref{section:case studies} outlines several simulated case studies, and Section~\ref{section:robot_experiments} showcases our framework on a physical multi-agent system. Finally, we conclude the study and discuss potential future work in Section~\ref{section:conclusion}.

\section{Preliminaries}
\label{section:preliminaries}

In this study, we use the General Reactivity of Rank 1 (GR(1)) fragment \cite{gr1} of Linear Temporal Logic (LTL) to specify desired tasks for each agent in a given environment. GR(1) synthesis is used to automatically generate correct-by-construction finite state machines (FSM). The generated strategy is implemented as a two-player game between the agent and the environment, where the FSM guarantees the agent satisfies the goal and safety specifications for any modeled environment action \cite{liu2013synthesis, kress2011correct}.

GR(1) allows for efficient synthesis while maintaining much of the expressiveness of LTL. In particular, GR(1) allows us to design temporal logic formulas ($\varphi$) with atomic propositions (AP) that can either be \textsf{True} ($\varphi \vee \neg\varphi$) or \textsf{False} ($\neg$\textsf{True}). With negation ($\neg$) and disjunction ($\vee$) one can also define the following operators: conjunction ($\wedge$), implication ($\Rightarrow$), and equivalence ($\Leftrightarrow$). There also exist temporal operators ``next" ($\bigcirc$), ``eventually" ($\Diamond$), and ``always" ($\square$). Further details of GR(1) can be found in \cite{gr1}. 

Our implementation uses the SLUGS reactive synthesis tool \cite{slugs}, which allows rules to be specified in a more human-intelligible structured slugs format using infix notation, non-negative integers, comparisons, and addition. These rules are automatically converted to GR(1) formulas  which are used to synthesize a reactive controller.

\section{Problem Formulation}
\label{section:problem formulation}

This paper studies a particular control synthesis problem where an agent in a heterogenous multi-agent team cannot complete its tasks because the environment has violated its assumptions at runtime. Formally, this occurs because an unmodeled environment behavior causes the specification to be unrealizable.

Let $\mathcal{P}$ denote the set of heterogeneous agents in a multi-agent team. When synthesizing a multi-agent controller, each agent $p\in \mathcal{P}$ within the team is given its own set of goal and safety specifications, denoted as $\varphi^p_o$ and $\varphi^p_s$, respectively. 

At the high-level, the environment is modeled using a coarse abstraction that divides the workspace into a set of $N$ discrete regions $\mathcal{S}=\{s_0, s_1, ..., s_{N-1}\}$. 
As low-level controllers are responsible for planning agent actions within each coarse region, they can be swapped in and out to accommodate different agent types without largely affecting the high-level actions. 

The set of known, irresolvable obstacles $\mathcal{O} \subset \mathcal{S}$ are accounted for as the set of safety specifications
\begin{equation}
\varphi^p_s := \bigwedge\limits_{s\in \mathcal{O}}\square \neg s.
\end{equation}

To account for the heterogeneity of the system, each agent $p\in \mathcal{P}$ is also modeled with an a priori known finite set of capabilities $C_p=\{c_{p0}, c_{p1}, c_{p2}, ...\}$. 
Examples of capabilities include ``open doors", ``inspect regions for hazards", or ``climb stairs". 

Obstacles that are resolvable but not known a priori are modeled as another subset $\mathcal{R} \subset \mathcal{S}$. Each resolvable obstacle $r\in \mathcal{R}$ has an associated action $c_r$ and set of states $S_r$ within which that action may be performed in order to resolve the obstacle and remove it from the environment. These properties are such that
\begin{equation}
c_r \in \bigcup_{p \in\mathcal{P}} C_p,\quad  S_r \subset (\mathcal{S}\setminus \mathcal{O}).
\end{equation}
Thus, an agent $p$ is considered to be capable of resolving an obstacle $r$ if $c_r \in C_p$. If an obstacle $r$ whose corresponding resolution action is not present in any agent's capability set, the obstacle $r$ will be considered as an unresolvable obstacle. By modelling agents and obstacles in this way, one can construct a robot team that is capable of resolving multiple types of obstacles by adding more robots with different capabilities to the team as needed. 

 It follows that any instance of an agent encountering an obstacle that it does not have the capability to resolve is considered a safety violation. We introduce an ``augmented" set of safety specifications $\varphi^p_a$, which contains the same specifications as $\varphi^p_s$ but in addition contains all of the additional safety specifications originating from unknown, resolvable obstacles:

\begin{equation}
\varphi^p_a := \varphi^p_s \bigwedge_{r \in R\ |\ c_r \notin C_p}\square \neg r 
\end{equation}
We can now formally define the problem statement.

\noindent\textbf{Problem Statement:} Assume a set of given controllers synthesized using $\varphi^p_s$, and a set of ``actual" environment specifications $\varphi^p_a$ such that one or more $\varphi^p_o$ are unrealizable under $\varphi^p_a$. Once the system is detected to violate $\varphi^p_a$ at runtime, create a generalizable formulation that assigns an agent $p$ to resolve and remove conflicting specifications in $\varphi^p_a$ such that the original synthesized controller satisfies $\varphi^p_a$.

\section{Controller Synthesis}
\label{section:controller synthesis}

To leverage the formal guarantees afforded by LTL, we synthesize navigation planners for each agent based on the planning framework detailed in~\cite{TowardsLoco}.  In this section, we provide an overview of the task and motion planners, which serve as the foundation that the proposed coordination layer will be built on. In subsequent sections, we augment the high-level navigation planning structure to further encode collaborative behaviors capable of resolving environment assumption violations at runtime, which constitutes the main contribution.

\begin{figure}[htb!]
\setlength{\belowcaptionskip}{-0.30cm}
\centerline{\includegraphics[width=0.5\textwidth]{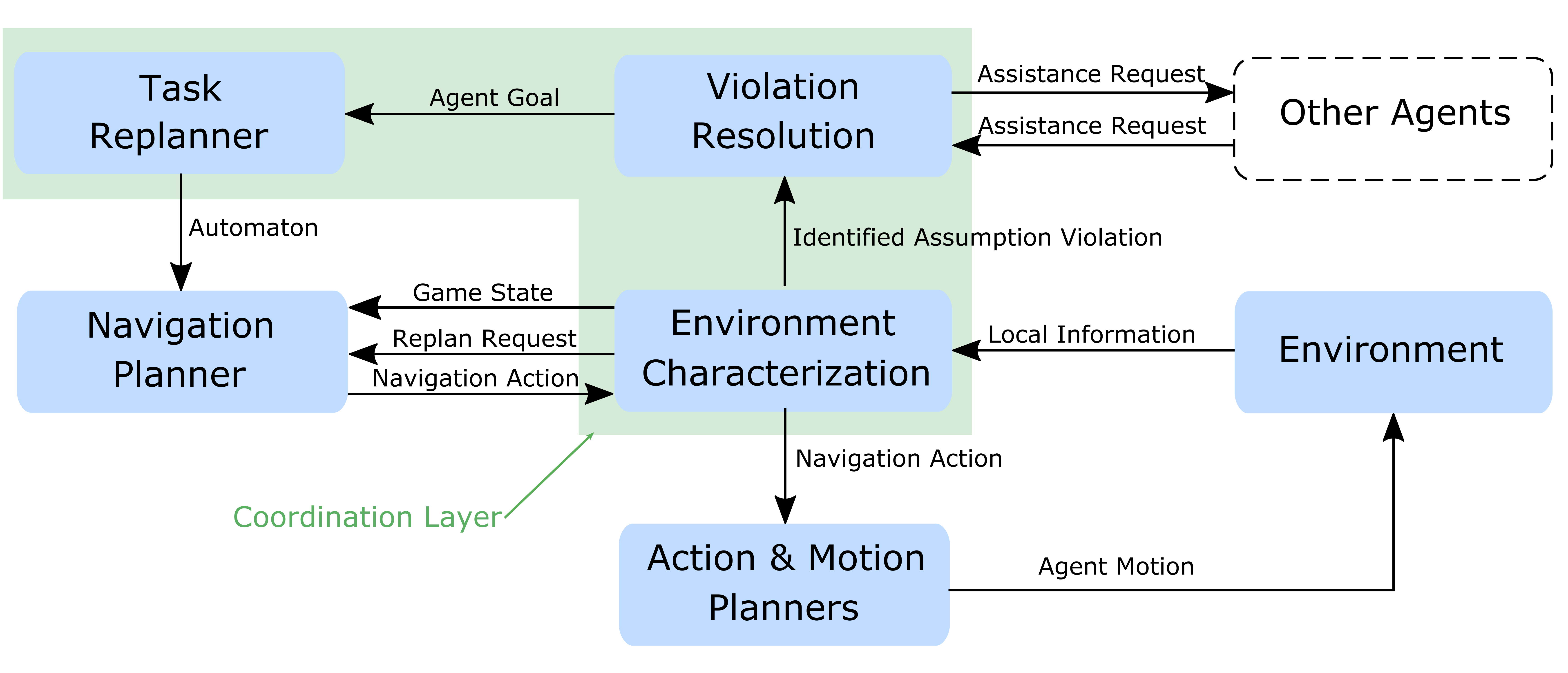}}
\caption{
Block diagram of collaborative task and motion planning framework. 
A coordination layer verifies that the desired actions generated by an offline-synthesized navigation planner are still safe based on environment information observed at runtime. If an environment assumption is violated, the navigation planner replans its current action to ensure the system enters a safe state while the coordination layer determines if the violation can be resolved by any other agent. The resolution is encoded and the plan progresses.}
\label{fig:frame}
\vspace{-0.15in}
\end{figure}

The approach from \cite{liu2013synthesis, TL_Planning, kress2011correct} adopted in~\cite{TowardsLoco} and used here is summarized as follows. To synthesize task planners, we construct two-player games between each agent and an abstracted environment.
We automatically encode a variety of propositions about how the game may evolve within LTL specifications, including initialization assumptions, environment safety assumptions, system safety guarantees, and system liveness properties. We synthesize an automaton, represented as a finite state machine (FSM), that guarantees the agent will always win the game as long as all the environment assumptions hold true at runtime. During runtime, the current environment state is input to the FSM, which outputs an action for the agent. Each action is guaranteed to meet the safety specifications while bringing the agent closer to completing its task.

In this work we synthesize planners for a bipedal robot Digit \cite{agility} and a quadcopter to study heterogeneous autonomous multi-agent navigation. We accomplish this by constructing a two-player game between each agent and its environment, where we treat all other agents as part of a partially observable environment and encode the possible moves each agent can expect the others to make in environment safety specifications. Thus, we can easily expand to more than two agents while keeping the planner synthesization a two-player game. As the planner prioritizes safety, the optimality of the controller is not necessarily guaranteed. For full details, see~\cite{TowardsLoco}.

\section{Coordination Layer Design}
\label{section:approach}
In this section, we introduce the coordination layer on top of the synthesized controllers, containing the elements enabling agents to identify new safety specifications at runtime, replan actions when necessary, identify and assign other agents to resolve obstacles, and adjust the behavior of each agent to execute the conflict resolution. Throughout, we assume the agents possess the necessary capabilities to facilitate the described sensory and communication actions. 
An overview of the proposed planning framework is in Figure~\ref{fig:frame}.

\subsection{Environment Characterization}
At runtime, the Environment Characterization (EC) block passes the game state abstracted from the environment to the Navigation Planner (NP) block and requests a navigation action. The EC block determines new safety specifications based on the observed environment and verifies if the navigation action would violate these specifications. This block is the main component responsible for observing the specifications in $\varphi^p_a$ that were not known during synthesis. 
If a safety violation occurs, the EC block signals the NP block to replan its current action.

\subsection{Safe Action Replanning}
The safe action replanning occurs within the NP block. At each step, the current environment state is fed to the FSM to generate a correct system action. When the replanning flag is raised, the navigation planner backtracks to the previous state in the FSM and extracts a new system action to avoid the safety violation. The new action is passed to the EC block, which passes it on to the lower-level planners if deemed safe.

\subsection{Violation Resolution}
When the EC block determines that the safety specifications based on the observed environment do not match the safety specifications used during offline synthesis, it also passes the details of the violation to the Violation Resolution (VR) block, including the action(s) $c_r$ required, as well as the state(s) $S_r$ at which those actions must be performed in to resolve the obstacle. The VR block then identifies which agent $p \in \mathcal{P}$ has the capability to resolve the obstacle, i.e., any agent $p$ such that $c_r \in C_p$. This component then broadcasts a request for an appropriate agent to assist. 

\subsection{Task Replanning}
The Task Replanner (TR) block receives incoming requests for assistance in the form of updated system goals and is responsible for adjusting the agent's strategy to assist the agent in need. 
Once a violation has been detected and assisting agents have been assigned, the TR block can trigger a resynthesis of each of the affected agent's controllers with their new objectives. The helping agent is assigned to resolving the obstacle, and the requesting agent is assigned to a known safety objective that does not interfere with the helping agent. After the controller detects that the obstacle has been resolved, another resynthesis is triggered, returning the agents to their original objectives. 

This method stores previous objectives in a stack; if a new resolvable obstacle is encountered while resolving a known obstacle, the resynthesis targets the new obstacle and pushes the previous set of objectives onto the stack. Once an obstacle is resolved, the top set of objectives in the stack is popped off, and controllers are resynthesized to these objectives. Beyond changing the target locations in the original objective, any number of new goal tasks or locations can be encoded in the new specifications to assist another agent that has encountered an assumption violation. Additionally, the agent that has encountered an obstacle that it cannot resolve on its own can be assigned a new task to complete while it waits for the obstacle to be resolved. A detailed pipeline of this process is shown in Algorithm~\ref{alg:resynthesis_alg}.

It should be noted that the algorithm is built upon an assumption of the existence of a sequence of obstacle resolutions that eventually results in the original objectives becoming achievable. In cases where this is not true, i.e., no agent exists that can resolve the obstacle or a set of resolution objectives are pushed onto the stack recursively, it would be impossible for the agents to accomplish their original set of objectives without external intervention. In such cases, synthesis or resynthesis of the automaton will fail, signaling the need for human operator intervention. One may avoid these cases by introducing agents with new capabilities.

\begin{algorithm}
\footnotesize
\SetAlgoLined
\For{$p \in \mathcal{P}$}{
$\varphi^p_o \leftarrow \text{initial objectives}$;\\
}
$\varphi_\text{stack} \leftarrow \emptyset$ ;\\
synthesize controllers;\\
\While{\rm system active}{
    execute controllers\;
    \If{\rm resolvable obstacle $r$ encountered by agent $p_r$ at $s_o$}{
    \uIf{$c_r \notin C_{p_r}$}{
    \text{push all of current $\varphi^p_o$ onto} $\varphi_\text{stack}$;\\
    change $\varphi^{p_r}_o$ to safe behavior;\\
    \For{$p \in \mathcal{P}$}{
        \uIf{$c_r \in C_p$}{
            \rm add state in $S_r$ to $\varphi^p_o$;\\
            \rm synthesize controllers;\\
            \rm break;\\
        }
    }
    \rm continue;\\
    }\ElseIf{r $\in \varphi^{p_r}_o$}{
    resolve obstacle $r$;\\
    \text{pop $\varphi^p_o$ off of} $\varphi_\text{stack}$;\\
    synthesize controllers;\\
    continue;\\
    }
    }
}
\caption{Resynthesis for Conflict Resolution}
\label{alg:resynthesis_alg}
\end{algorithm}

Notice that since the controllers are synthesized by GR(1) reactive synthesis, they are correct by construction \cite{liu2013synthesis}.

We additionally develop an alternative method involving an additional runtime-assignable goal location specified in the assisting agent's liveness specifications, which enables a limited amount of coordination without resynthesis. However, details of this method are omitted due to space constraints. Instead, for full implementations of either method, we point the reader to the GitHub repository located at \url{https://github.com/GTLIDAR/safe-nav-locomotion}.

\section{Case Studies}
\label{section:case studies}

In this section, we implement and evaluate the framework described in Section~\ref{section:controller synthesis}, which was primarily built to synthesize controllers for the bipedal walking robot platform Digit, designed by Agility Robotics \cite{agility}. We consider a second quadcopter agent in the environment that has dramatically different capabilities in both mobility and manipulation. A quadcopter, which lacks the ability to manipulate objects, is instead able to perform maneuvers that are unavailable to Digit, such as backward movement or $180^\circ$ turns in a single region. In this study, the quadcopter is assumed to fly above Digit at all times, so that collision is not a concern. This assumption is made to reduce the complexity of the controller synthesis for these case studies, as employing collision avoidance, while trivial from an implementation standpoint thanks to~\cite{TowardsLoco}, drastically increases the computation time. Additionally, we preserve the belief space planning framework proposed in \cite{TowardsLoco, bharadwaj2018synthesis}, allowing Digit to infer the quadcopter location if it is out of Digit's visible range. Digit's locomotion planner is designed based on the phase-space planning framework in \cite{zhao2017robust}.

Resolvable obstacles are also implemented within the simulated environment. We recall from Section~\ref{section:problem formulation} that each resolvable obstacle $r$ has an associated state $S_r$ where an action $c_r$ must be performed by an agent $p$ for which $c_r \in C_p$ in order to resolve the obstacle and remove it from the environment. These properties are directly inserted into the simulation environment. The set of resolvable obstacles $\mathcal{R}$ and the set of agents and their capabilities $C_p$ are implemented as separate dictionary data structures. As such, this framework is generalizable 
as one would simply need to add the appropriate agent capabilities and obstacle resolutions to each dictionary.

To represent the potential for obstacles to appear at runtime, two separate simulated environments are initialized, one without the resolvable obstacle. This instance of the environment, representing $\varphi^p_s$, is used for the initial synthesis, and then the resulting controller is applied to the environment instance containing the resolvable obstacles, which corresponds to $\varphi^p_{a}$. At runtime, if an agent enters a state containing a resolvable obstacle $r$ that it is unable to resolve (i.e., violates $\varphi^p_{a}$), the controller is able to check that a violation has happened in the simulated environment, and sends this information to the simulation to assign new objectives to each agent accordingly, such that the agent $p$ tasked with resolving the obstacle fulfills $c_r \in C_p$. Thus, the simulation running each of the controllers is responsible for the VR and TR blocks of the coordination layer, while the separate simulated environment instances simulate the EC component. The Safe Action Replanner is built directly into the NP block.

Three case studies utilizing the synthesized controllers are presented to evaluate the proposed approach. For each case study, an environment is created where a quadcopter and Digit are each running on their own controller and have their own task objectives to complete. The environment is abstracted into a 7$\times$13 coarse set of regions $\mathcal{S}=\{s_0, s_1, ..., s_{90}\}$ such that $s_0$ is the northwestern-most region and increments following English reading orientation (i.e. incrementing left to right, then starting at the leftmost region on the next row). This setup can be seen in Figures~\ref{fig:open_casestudy}-\ref{fig:sense_casestudy2}.

We consider a team of agents $P = \{\tt{quadcopter}$ and $\tt{Digit}\}$ with unique capabilities
$C_\text{quad} := \{\tt{sense}\}$, $C_\text{Digit} := \{\tt{push}\}$.
Resolvable obstacles that may appear within a region $s_i$ in the environment consist of two types: $r=\tt{door}$ and $r=\tt{uncertainty}$. A resolvable obstacle of type $r=\tt{door}$, if found in $s_i$, has properties
\begin{align}
    S_r = \{s_i\}, \;
    c_r = \tt{push},
\end{align}
and represents physical doors that the quadcopter cannot fly through, but are able to be opened by Digit. Resolvable obstacles of type $r=\tt{uncertainty}$ have the properties
\begin{align}
    S_r = \{s_{\rm north}, s_{\rm east}, s_{\rm south}, s_{\rm west}\}, \;
    c_r = \tt{sense},
\end{align}
and represent regions in which Digit is uncertain about its capabilities to safely traverse through the environment, but the quadcopter is able to scout them by visiting any adjacent region. 

We design a set of objective specifications for each agent $p \in \mathcal{P}$ such that the agent alternates between visiting two regions in the environment $s_\text{A}, s_\text{B} \in \mathcal{S}$, with an AP $\tt{scout}$ unique to each agent, which initializes to \textsf{False}, to track which region the agent should head towards. The set of objective specifications is thus
\begin{align}
    \label{eq:patrol_specs}\text{Patrol}_p&(s_\text{A}, s_\text{B}) = \\ \nonumber&\square\Diamond(s_\text{A} \wedge \neg {\tt scout}_p) \\
    \nonumber\wedge&\square((s_\text{A} \wedge \neg {\tt scout}_p) \Rightarrow \bigcirc ({\tt scout}_p \wedge \neg s_\text{A})) \\
    \nonumber\wedge&\square\Diamond(s_\text{B} \wedge {\tt scout}_p) \\
    \nonumber\wedge&\square((s_\text{B} \wedge {\tt scout}_p) \Rightarrow \bigcirc (\neg {\tt scout}_p \wedge \neg s_\text{B})) \\
    \nonumber\wedge&\square\neg(s_\text{A} \wedge {\tt scout}_p)\\
    \nonumber\wedge&\square\neg(s_\text{B} \wedge \neg {\tt scout}_p).
\end{align}

While the figures in this section mainly feature abstractions of the environment in order to easily illustrate the behaviors of each agent, the computed control actions are applicable to a real 3D simulation environment, as shown in Figure~\ref{fig:Sim}. Those two subfigures show the real-world interactions resulting from the behavior in the first case study.
\subsection{Case Study 1: Opening A Door}
\begin{figure*}[]
     \centering
     \begin{subfigure}[t]{\textwidth}
         \centering
         \nonumber
         \includegraphics[width=0.8\textwidth]{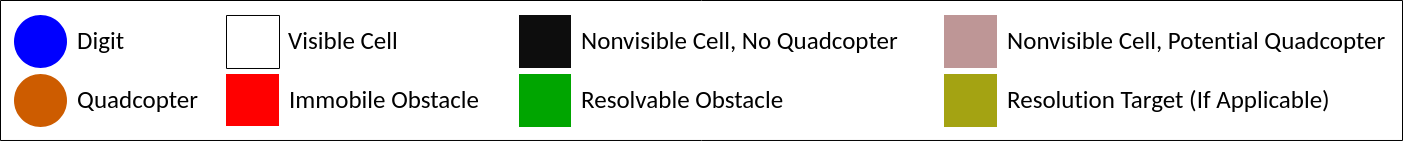}
         \label{fig:legend}
     \end{subfigure}
     \hfill
     
     \begin{subfigure}[t]{0.3\textwidth}
         \centering
         \includegraphics[width=0.8\textwidth]{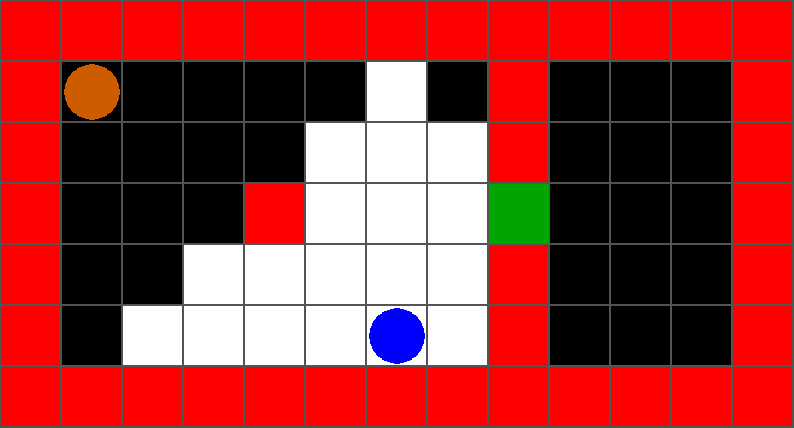}
         \caption{Initial Configuration}
         \label{fig:open_0}
     \end{subfigure}
     \hfill
     \begin{subfigure}[t]{0.3\textwidth}
         \centering
         \includegraphics[width=0.8\textwidth]{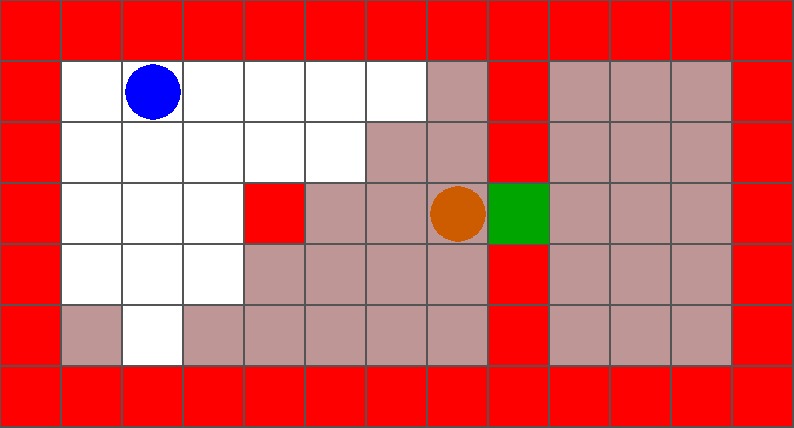}
         \caption{The quadcopter's controller encounters a door, so instead it elects to have the quadcopter wait in the preceding region.}
         \label{fig:open_1}
     \end{subfigure}
     \hfill
     \begin{subfigure}[t]{0.3\textwidth}
         \centering
         \includegraphics[width=0.8\textwidth]{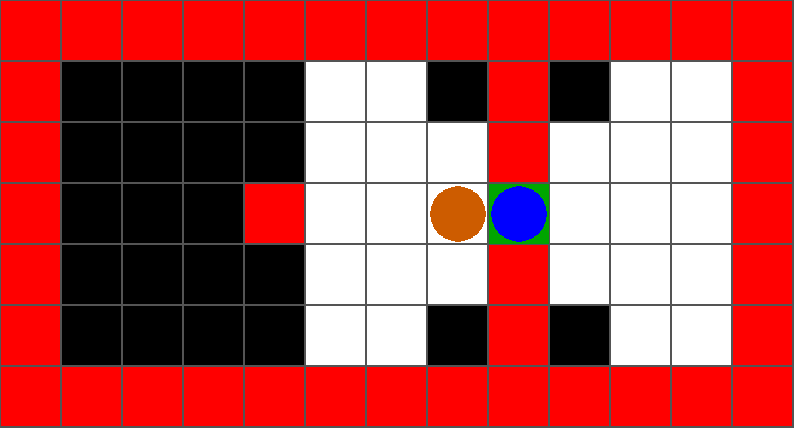}
         \caption{Digit opens the door and the original objectives are reinstated. Both agents are now able to complete their objectives.}
         \label{fig:open_2}
     \end{subfigure}
     \setlength{\belowcaptionskip}{-0.80cm}
        \caption{Execution of case study 1 leveraging Digit's manipulation abilities to clear the path for the quadcopter. Digit's objective is patrolling the left room, while the quadcopter's objective is delivering to the right room. However, the quadcopter discovers a closed door separating the two rooms at runtime, prompting Digit to navigate to open it so that both agents are able to complete their objectives.}
        \label{fig:open_casestudy}
\end{figure*}

The first case study leverages Digit's manipulation capability in the environment with higher dexterity and power than the quadcopter. The quadcopter is tasked with patrolling between $s_{\rm homeQ}$ and $s_{\rm awayQ}$, where $s_{\rm homeQ}$ is a region in the left room and $s_{\rm awayQ}$ is in the right room. Digit is tasked with patrolling between $s_{\rm homeC}$ and $s_{\rm awayC}$, where both $s_{\rm homeC}$ and $s_{\rm awayC}$ are in the left room:
\begin{align}
    \varphi^{\text{quad}}_o &:= \text{Patrol}_\text{quad}(s_{\rm homeQ}, s_{\rm awayQ}),\\
    \nonumber\varphi^{\text{Digit}}_o &:= \text{Patrol}_\text{Digit}(s_{\rm homeC}, s_{\rm awayC}).
\end{align}

At runtime, the quadcopter discovers an obstruction at $s_{\rm door} = s_{47}$ while in $s_{\rm safe} = s_{46}$ that prevents it from accomplishing its objective in the form of a closed door that cannot be flown through but can be opened by Digit. A resynthesis of objectives is triggered, where the quadcopter is now tasked with hovering outside the door, and Digit is tasked with visiting one of its initial patrol points and the closed door, resulting in \begin{align}\varphi^{\text{quad}}_o &:= \text{Patrol}_\text{quad}(s_{\rm safe}, s_{\rm safe}),\\ 
    \nonumber\varphi^{\text{Digit}}_o &:= \text{Patrol}_\text{Digit}(s_{\rm homeC}, s_{\rm door}).\end{align}

Once Digit visits the door, it is considered open and the obstacle is removed from the environment, triggering another resynthesis which returns both agents to their original target objectives. A walkthrough of the execution of this case study is shown in Figure~\ref{fig:open_casestudy}. For this case study, we also used low-level planners to generate safe motions for the quadcopter and Digit, including center of mass (CoM) trajectories and foot placements. A 3D visualization can be seen in Figure~\ref{fig:Sim} and the included video.

\subsection{Case Study 2: Scouting Ahead}
The second case study involves Digit encountering several states and not knowing whether each state is safe to traverse on foot, requiring the help of the quadcopter's heightened sensing capabilities. To this end, the quadcopter is set to patrol between $s_{\rm homeQ}$ and $s_{\rm awayQ}$, where $s_{\rm homeQ}$ and $s_{\rm awayQ}$ are in the left room, while Digit must patrol between $s_{\rm homeC}$ and $s_{\rm awayC}$, where $s_{\rm homeC}$ is a region in the left room and $s_{\rm awayC}$ is in the right room:
\begin{align}
    \label{eq:scout_init}
    \varphi^{\text{quad}}_o &:= \text{Patrol}_\text{quad}(s_{\rm homeQ}, s_{\rm awayQ}),\\
    \nonumber\varphi^{\text{Digit}}_o &:= \text{Patrol}_\text{Digit}(s_{\rm homeC}, s_{\rm awayC}).
\end{align}

At runtime, Digit encounters region $s_{\rm uncertain1} = s_{34}$ while at $s_{\rm safe1} = s_{33}$, and it is unsure about its ability to traverse this region. A resynthesis is triggered, where the quadcopter is tasked with observing the uncertain region by visiting any of the adjacent regions (in this case, we select the region $s_{\rm uncertain1W}$ directly west of $s_{\rm uncertain1}$), resulting in 
\begin{align}    \varphi^{\text{quad}}_o &:= \text{Patrol}_\text{quad}(s_{\rm homeQ}, s_{\rm uncertain1W}),\\ 
    \nonumber\varphi^{\text{Digit}}_o &:= \text{Patrol}_\text{Digit}(s_{\rm safe1}, s_{\rm safe1}).
\end{align}

Once the quadcopter observes the unknown region, the resolvable obstacle is removed from the environment. If the quadcopter senses that the region is not traversable by Digit, then the region is added to $O$ and will be considered as an immovable obstacle during future synthesis. In this specific case study, $s_{\rm uncertain1W}$ is found to be untraversable. 

\begin{figure*}[t]
\setlength{\belowcaptionskip}{-0.30cm}
     \hfill
     \begin{subfigure}[t]{0.3\textwidth}
         \centering
         \includegraphics[width=\textwidth]{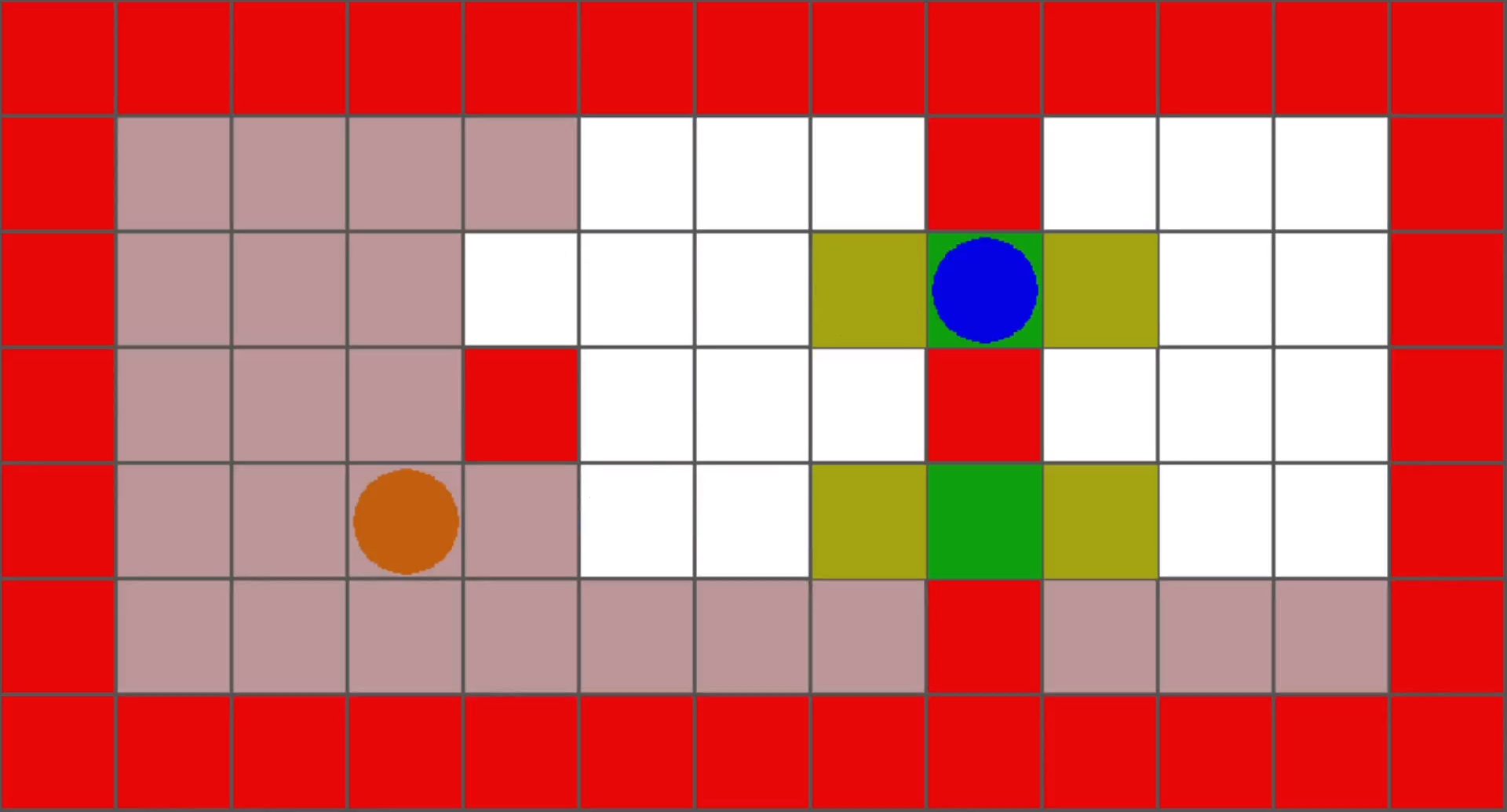}
         \caption{Whenever Digit's controller encounters an uncertain region, it has Digit wait in the preceding region.}
         \label{fig:sense_1}
     \end{subfigure}
     \hfill
     \begin{subfigure}[t]{0.3\textwidth}
         \centering
         \includegraphics[width=\textwidth]{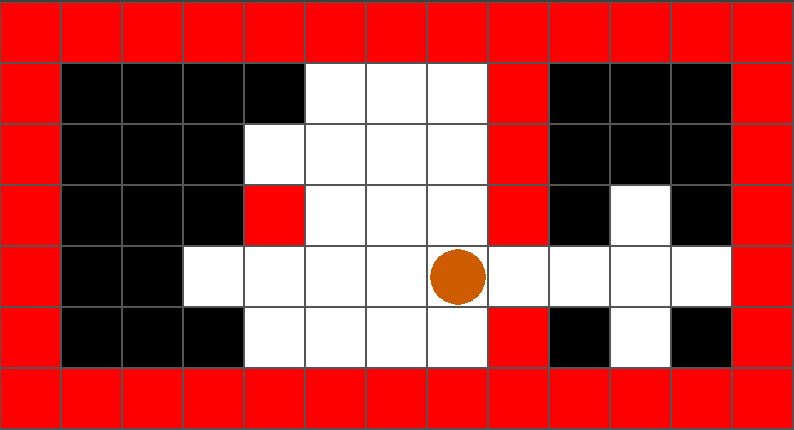}
         \caption{The quadcopter senses that the first region is an obstacle, but the second region is traversable by Digit.}
         \label{fig:sense_4}
     \end{subfigure}
     \hfill
     \begin{subfigure}[t]{0.3\textwidth}
         \centering
         \includegraphics[width=\textwidth]{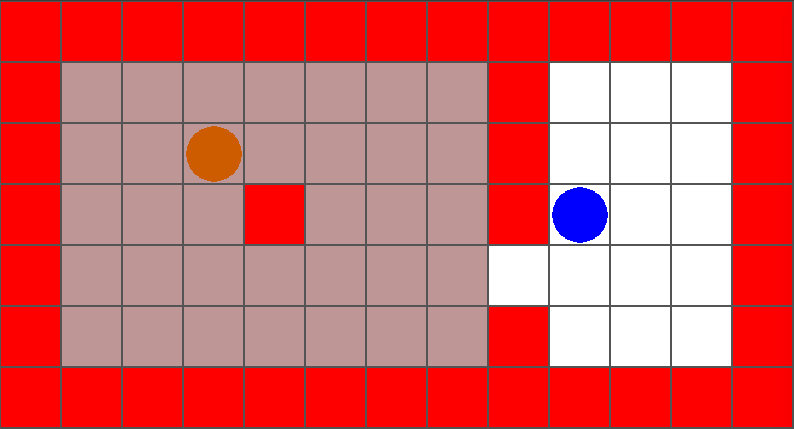}
         \caption{Original objectives are reinstated, with both agents now able to meet their objectives.}
         \label{fig:sense_5}
     \end{subfigure}
     \setlength{\belowcaptionskip}{-0.50cm}
        \caption{Execution of case study 2 leveraging the quadcopter's sensory capabilities to find a traversable path for Digit. The quadcopter's objective is to patrol the left room, while Digit's objective is to deliver something to the right room. However, Digit encounters several uncertain regions, requiring the quadcopter to observe each until it finds a traversable region.}
        \label{fig:sense_casestudy2}
\end{figure*}

The two agents are returned to their original objectives, outlined in~\eqref{eq:scout_init}, before Digit encounters another uncertain state at $s_{\rm uncertain2} = s_{60}$ while in $s_{\rm safe2} = s_{59}$, where the process repeats. The quadcopter is tasked with visiting $s_{\rm uncertain2W}$, directly west of $s_{\rm uncertain2}$, while Digit is instructed to stay at $s_{\rm safe2}$. The uncertain region is found to be traversable by Digit, and both agents are returned to their original objectives, now able to fulfill them. A walkthrough of the execution of this case study is shown in Figure~\ref{fig:sense_casestudy2}.

\subsection{Case Study 3: Chain of Conflicts}
\label{sec:case3}
The third case study merges the previous two and requires both agents to resolve an obstacle. For this case study, the quadcopter encounters a door while on its way to resolving an uncertain region encountered by Digit, requiring Digit to first open the door, thus demonstrating the coordination layer's ability to handle multiple resolvable obstacles in a chain when required. 

Initially, Digit is tasked with patrolling between $s_{\rm homeC}$ in the leftmost room and $s_{\rm awayC}$ in the center room, while the quadcopter is tasked with patrolling between $s_{\rm homeQ}$ and $s_{\rm awayQ}$, both in the rightmost room:
\begin{align}
    \varphi^{\text{quad}}_o &:= \text{Patrol}_\text{quad}(s_{\rm homeQ}, s_{\rm awayQ}),\\
    \nonumber\varphi^{\text{Digit}}_o &:= \text{Patrol}_\text{Digit}(s_{\rm homeC}, s_{\rm awayC})
\end{align}

Digit encounters an uncertain state at $s_{\rm uncertain} = s_{43}$ while in $s_{\rm safeC} = s_{43}$, triggering a resynthesis requiring the quadcopter to sense the true traversibility of that state by visiting $s_{\rm uncertainE} = s_{44}$, resulting in 
\begin{align}
    \varphi^{\text{quad}}_o &:= \text{Patrol}_\text{quad}(s_{\rm homeQ}, s_{\rm uncertainE}),\\ 
    \nonumber\varphi^{\text{Digit}}_o &:= \text{Patrol}_\text{Digit}(s_{\rm safeC}, s_{\rm safeC}).
\end{align}

However, the quadcopter encounters a closed door at $s_{\rm door} = s_{47}$ while in $s_{\rm safeQ} = s_{48}$ on its way to resolve Digit's uncertainty, triggering another resynthesis where Digit is tasked with opening the door, resulting in 
$
    \varphi^{\text{quad}}_o := \text{Patrol}_\text{quad}(s_{\rm safeQ}, s_{\rm safeQ}), \ 
    \nonumber\varphi^{\text{Digit}}_o := \text{Patrol}_\text{Digit}(s_{\rm safeC}, s_{\rm door}).
$
Once the door is opened and resolved, the quadcopter travels to the uncertain state and resolves that obstacle, resulting in both agents being able to accomplish their objectives.  

For this case study, in lieu of a figure illustrating the abstracted simulation, we apply the computed control actions to a physical robotic system consisting of a bipedal robot and a quadcopter, detailed in the next section.

\section{Hardware Implementation}
\label{section:robot_experiments}
We replicated case study 3 (see Sec.~\ref{sec:case3}) on physical hardware, which involved coordination between a bipedal robot Digit and a quadcopter Parrot Bebop 2. A video of the experiment is included as supplementary material.

\begin{figure}[h]
\setlength{\belowcaptionskip}{-0.60cm}
\centering
\begin{subfigure}{.4\textwidth}
  \centering
  \includegraphics[width=0.8\linewidth]{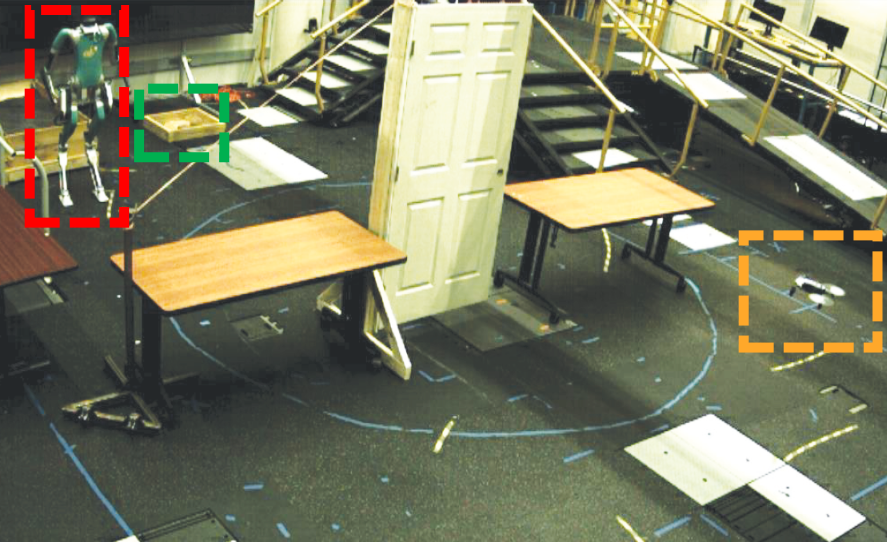}

  \label{fig:experimentSetup}
\end{subfigure}%
\hfill
\begin{subfigure}{.4\textwidth}
  \centering
  \includegraphics[width=0.8\linewidth]{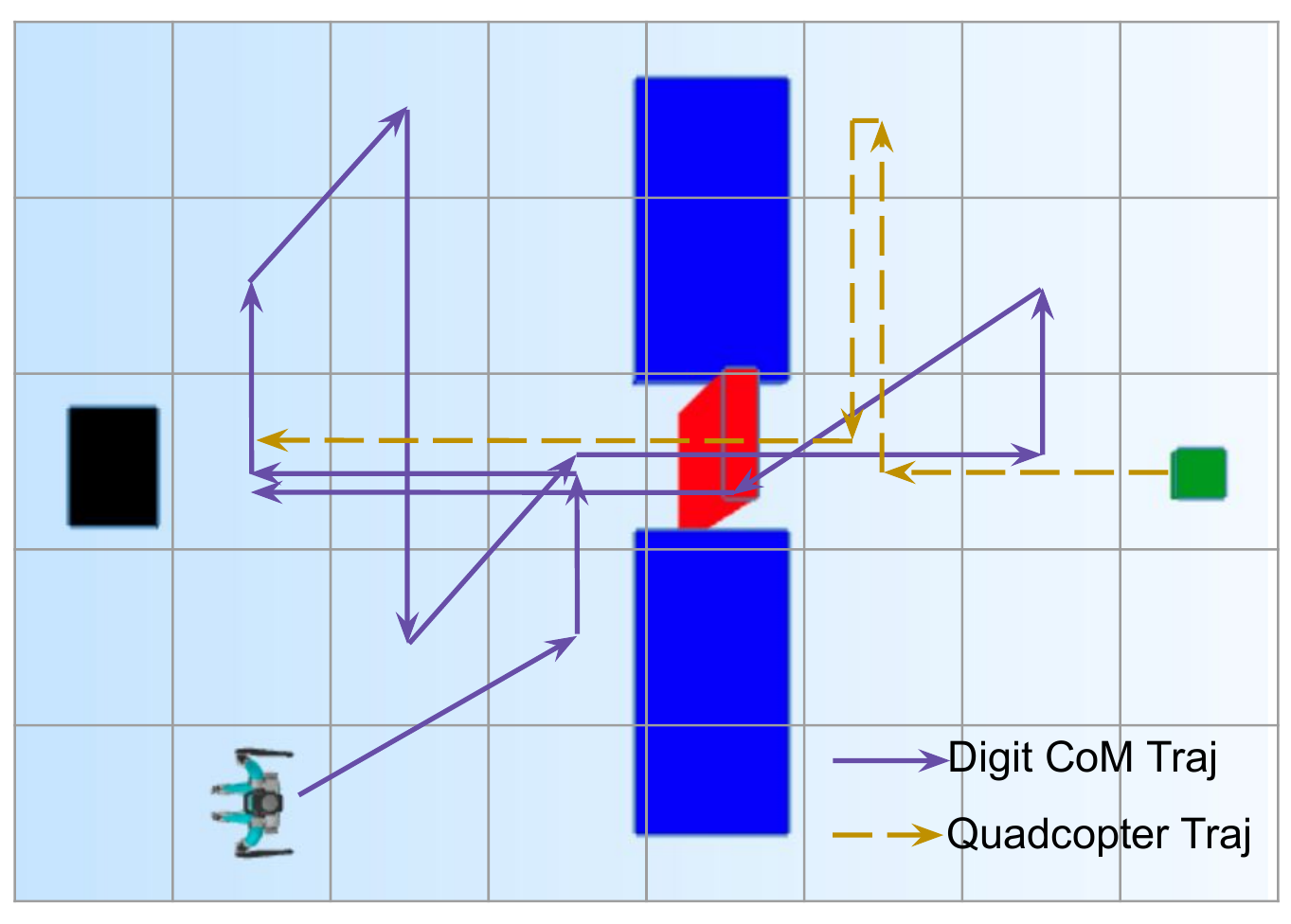}
  \label{fig:experimentTraj}

\end{subfigure}
\caption{Hardware experiment Setup. In the top figure, the desks represent walls --- static obstacles in the LTL setup, and the door represents the resolvable obstacle. The tray boxed in green represents the uncertain region. The robot boxed in red is the Digit bipedal robot and the one boxed in blue is the Bebop quadcopter at initial positions. In the bottom figure, the robots' motion trajectories are represented via arrowed lines in a similar simulated environment. }
\label{fig:doorOpen}
\end{figure}

Digit has an upper body that consists of a torso and two arms with 4 actuatable joints on each, which allows it to achieve manipulation skills and enables it to open a door during the hardware experiment (see Fig.~\ref{fig:doorOpen}). For this experiment, we use the Digit controller provided by Agility Robotics, which takes the 2D world waypoint position as input and automatically plans Digit's walking gait to approach the desired position. Hence, at runtime, Digit is commanded to track the waypoints sent by the task and motion planner to fulfill the patrolling task. When the new task of opening a door is triggered, Digit navigates to the door and opens it by pushing the door with its left arm.

Localization of the quadcopter is performed with an overhead Vicon motion capture system (servo rate 20 Hz). The LTL framework computes setpoints for the quadcopter to track, and tracking is performed using a quadratic programming based controller, also operating at 20 Hz. 

The experiment was set up to simulate a disaster relief scenario where the two robots collaborate in a disaster-struck house. Two desks (simulating walls) and a door are placed inside the Vicon motion capture area, with the desks as static obstacles and the door as the resolvable obstacle (i.e., can be opened). 
The uncertain region is represented using a tray with bricks (boxed in green in Fig. \ref{fig:doorOpen}). Hence, Digit is tasked with patrolling between $s_{\rm uncertain} = s_{43}$ and $s_{\rm awayC}$ in the center room.

During the hardware experiment, new goals are extracted from the automaton synthesized offline. Since commands extracted from the automaton take negligible time, the FSM can generate control policies at runtime. The hardware experiments showed that the proposed LTL framework was able to command the robot team to successfully open the door, sense the uncertain region, and fulfill the original tasks. The experiment was recorded in the video attachment to this submission.
 
The hardware implementation reveals a few new challenges.
First, the synchronization between the two different robots makes the collaborative task less efficient as they reach the setpoints at different times, which results in longer time for the task completion.
The initial LTL synthesis took around one minute, leaving Digit standing, and the drone airborne and exhausting their battery. Efforts can be made to reduce the initial synthesis time. 
Additionally, incorporating a high-performance whole-body motion controller for Digit such as~\cite{Aziz2021WBC} would enable improved walking performance, though that is beyond the scope of this work.

\section{Conclusion and Future Work}
\label{section:conclusion}
We presented a generalizable approach to identifying and resolving environment assumption violations discovered at runtime by automatically leveraging the capability of heterogeneous agents. This allows the team of agents to recover from cases where their objectives become unrealizable due to runtime-observed violations. We implemented this approach in a grid world simulation and generated safe 3D CoM motion plans for a bipedal robot and a quadcopter.
Future directions of research include 
implementing deadlock resolution strategies and further developing the non-resynthesis solution to enable more complex runtime-assignable behaviors.
Additionally, more complex sets of robot teams can be constructed to resolve a wider range of obstacles using the proposed approach. Future work can be done to catalogue a comprehensive library of robots, capabilities, and obstacle resolutions to enable broader applications, such as multi-quadcopter teaming to deliver a battery for Digit charging and executing long-duration navigation for package delivery. Robots that can pull doors open and remove debris can be added to the robot team to form a versatile team that aims for search and rescue applications. Migrating towards signal temporal logic is also of future research interest so that specifications about resolution time can be included, enabling the proposed approach to be deployed in timing-critical missions.

{
\printbibliography
}
\end{document}